\documentclass[11pt]{article}

\usepackage[preprint]{acl}

\usepackage{times}
\usepackage{latexsym}

\usepackage[T1]{fontenc}

\usepackage[utf8]{inputenc}

\usepackage{microtype}

\usepackage{inconsolata}

\usepackage{graphicx}

\usepackage{subfiles}
\usepackage{graphicx}
\usepackage{makecell}

\usepackage{tcolorbox} 
\newtcolorbox{fullwidthbox}{
    colback=gray!10, 
    colframe=black, 
    width=\textwidth, 
    sharp corners, 
    boxrule=1pt, 
    left=5pt, right=5pt, top=5pt, bottom=5pt 
}

\title{Measuring Distribution Shift in User Prompts \\ and Its Effects on LLM Performance}

\author{Parker Seegmiller \and Sarah Masud Preum \\ Department of Computer Science \\
        Dartmouth College \\ Hanover, NH, USA \\ \texttt{\{pkseeg.gr, Sarah.Masud.Preum\}@dartmouth.edu}}

\begin{document}
\maketitle

\begin{abstract}
    LLMs are increasingly deployed in dynamic, real-world settings, where the distribution of user prompts can shift substantially over time as new tasks, prompts, and users are introduced to a deployed model. Such \textbf{natural prompt distribution shift} poses a major challenge to LLM reliability, particularly for specialized models designed for narrow domains or user populations. Despite attention to out-of-distribution robustness, there is very limited exploration of \textbf{measuring} natural prompt distribution shift in prior work, and its impact on deployed LLMs remains poorly understood. We introduce the \textbf{L}LM \textbf{E}valuation under \textbf{N}atural prompt \textbf{S}hift (\textbf{LENS}) framework: a data-centric approach for quantifying natural prompt distribution shift and evaluating its effect on the performance of deployed LLMs. We perform a large-scale evaluation using 192 real-world post-deployment prompt shift settings over time, user group, and geographic axes, training a total of 81 models on 4.68M training prompts, and evaluating on 57.6k prompts. We find that even moderate shifts in user prompt behavior correspond with large performance drops (73\% average loss) in deployed LLMs. This performance degradation is particularly prevalent when users from different latent groups and geographic regions interact with models and is correlated with natural prompt distribution shift over time. We systematically characterize how \textbf{LLM instruction following ability degrades over time and between user groups}. Our findings highlight the critical need for data-driven monitoring to ensure LLM performance remains stable across diverse and evolving user populations.
\end{abstract}

\section{Introduction}

\begin{figure}
    \centering
    \includegraphics[width=0.5\textwidth]{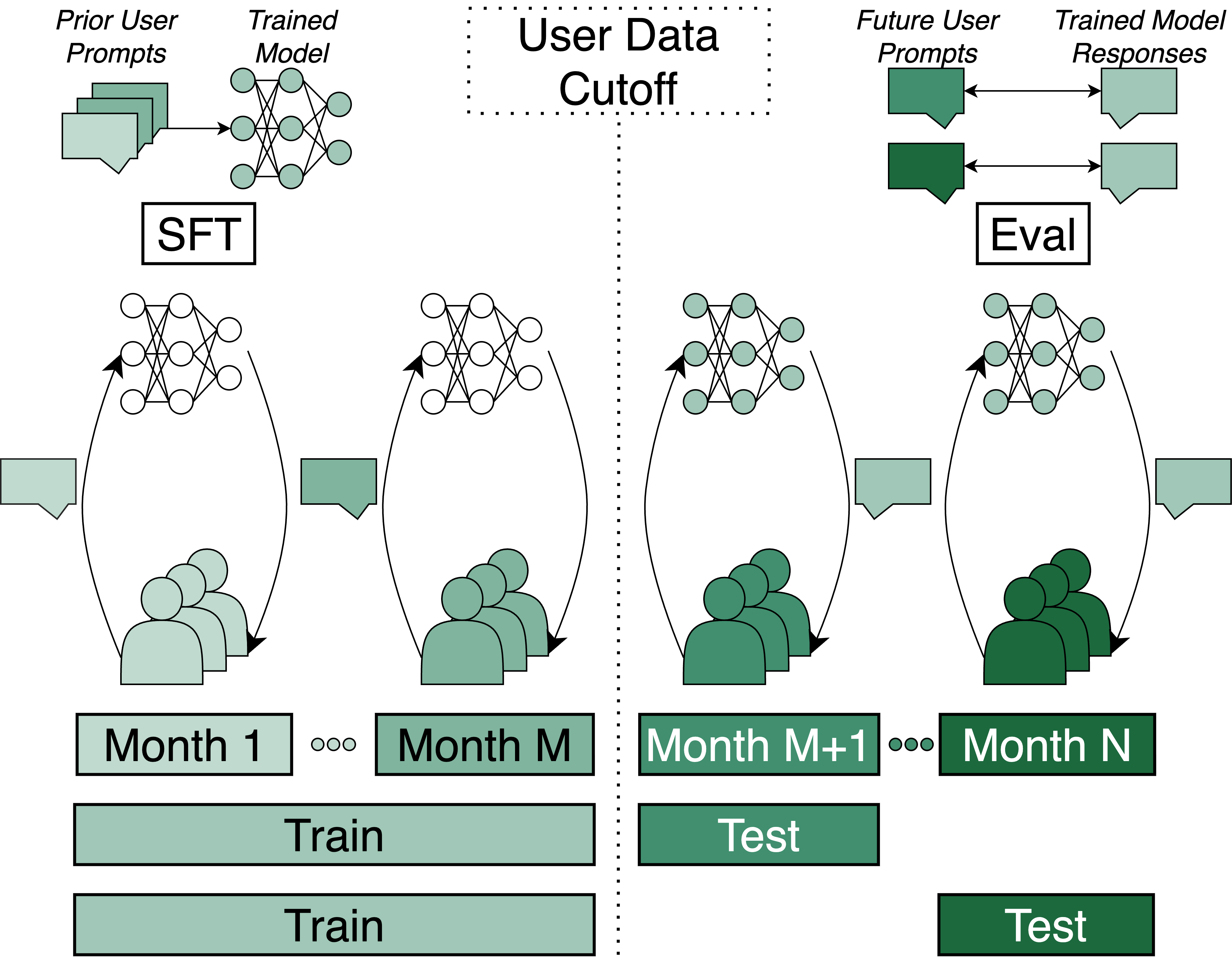}
    \caption{LLM user dynamics change considerably over time, affecting prompt distributions and performance of models trained on prior user prompts. We measure these natural prompt distribution shifts and study their effects on LLM performance.}
    \label{fig:preview}
\end{figure}

The way humans interact with large language models (LLMs) evolves naturally over time, location, and user groups \cite{ma2024death, tafreshipour2025prompting, seegmiller2024llms}. This covariate distribution shift can substantially affect model reliability and safety \cite{yuan2023revisiting, myntti2025register, lazaridou2021mind}, thereby affecting user satisfaction \cite{lin2024interpretable} and overall model performance \cite{yuan2023revisiting, yang2023glue}. We focus on a specific type of covariate shift, which we refer to as \textbf{natural prompt distribution shift}: measurable change in the natural distribution of user prompts encountered by a model after deployment, relative to the distribution of user prompts seen during instruction fine-tuning. 

\begin{figure*}
    \centering
    \includegraphics[width=\textwidth]{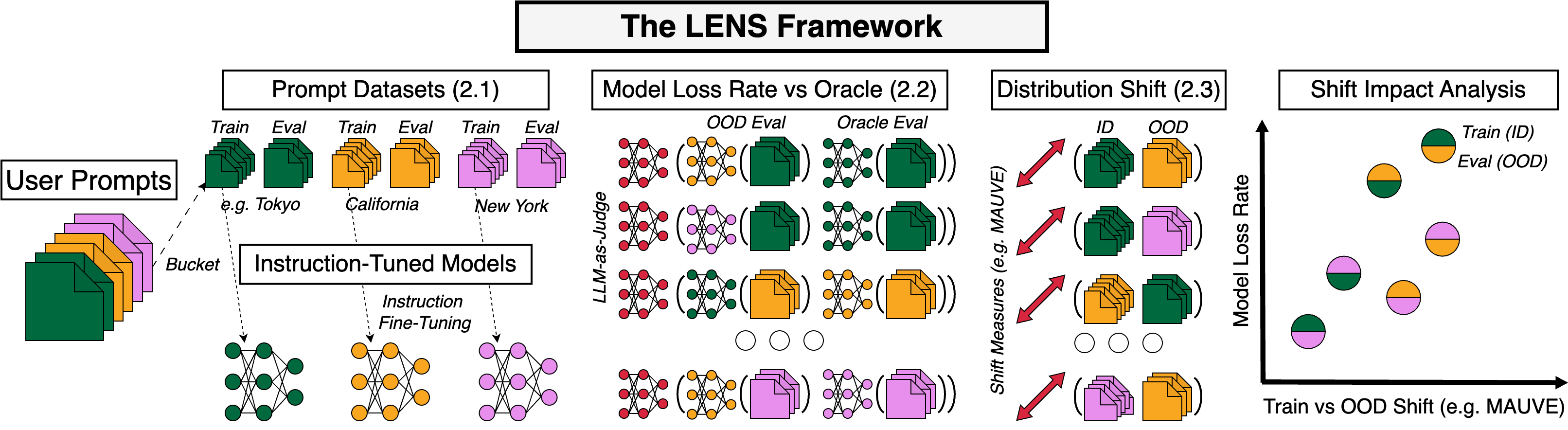}
    \caption{The LENS framework for investigating the relationship between instruction tuning dataset distributions and user prompt distributions in realistic deployment settings. LENS creates \{ID/OOD\} settings over time, between user groups, and across geographic regions (Section \ref{sec:framework_datasets}). LLM OOD performance (Section \ref{sec:framework_training}) is compared with unsupervised measures of distributional shift (Section \ref{sec:framework_measures}) between prompts seen during instruction fine-tuning vs evaluation. LENS enables analysis of the impact of distributional shift on LLM performance.}
    \label{fig:framework}
\end{figure*}


Real-world user prompts are often repurposed as instruction tuning data, as LLMs fine-tuned on such prompts and collected responses have demonstrated strong performance in downstream deployments \cite{olmo20242, ivison2023camels, cui2025comprehensive}.  However, as with any inputs into a deployed ML system, these prompts are subject to distributional shift from training to deployment (see Figure \ref{fig:preview}) \cite{jang2024driftwatch, luu2022time}. This is particularly prevalent for smaller LLMs with dedicated user bases in deployment, such as those deployed for customer service \cite{xu2024retrieval} or medical dialogue \cite{garcia2024medical}, as dynamics such as the introduction of new users may affect the distribution of user prompts. If LLMs are not robust to such natural shifts in user prompts after deployment, this can have disastrous effects on user satisfaction and  model reliability \cite{massenon2025my, wang2024understanding}.

Despite the practical significance of natural prompt distribution shift and its effects, robust measurement and analysis of the impact of such shifts on deployed LLM performance remain open challenges. Evaluation of LLMs under prompt distribution shift can be viewed as out-of-distribution (OOD) robustness, a well-studied challenge in machine learning \cite{hendrycks2021many, hendrycks2020pretrained}. There is some evidence that LLMs fail to generalize across domains in specific tasks \cite{yuan2023revisiting, wenzel2022assaying, teney2023id, wu2025pay}. For example, \citet{wu2025natural} show that natural semantic evolution of context paragraphs from pretraining data may lead to up to 70\% decline in LLM reading comprehension performance. Similarly, \citet{yang2023glue} show that LLMs frequently experience 20\% performance drop in OOD evaluations on GLUE benchmark tasks.

However, evaluating LLM OOD robustness to ``in the wild'' user prompt distribution shift has yet to be studied at scale. This is a challenge for two reasons. First, distribution shift is not trivial to define or measure for massive-scale corpora of natural language such as user prompts. In prior evaluations of LLM robustness, distribution shift is implicitly assumed rather than measured \cite{sun2024evaluating, Hupkes2022ATA, yuan2023revisiting, Gupta2023WhispersOD, teney2023id}. Second, the extent to which intentional transfer learning from rounds of pre-training mitigates post-trained LLM performance degradation due to semantic and stylistic divergence between training prompts and those seen in deployment is an open question \cite{zhao2024deciphering, jia2024distribution}. Robust measurement of natural prompt distribution shifts, and controlled analysis of their impact on LLM performance, will enable a deeper understanding of real-world dynamics affecting LLM evaluation and help to develop more robust model performance \textit{in the wild}. 

To address these open challenges, we study two core research questions in this paper. RQ1: How much prompt distribution shift can be observed in the wild? RQ2: How do natural prompt distribution shifts affect performance of pre-trained LLMs?

We introduce the \textbf{L}LM \textbf{E}valuation under \textbf{N}atural prompt distribution \textbf{S}hift (\textbf{LENS}) framework for analyzing the relationship between instruction tuning dataset distributions and user prompt distributions in deployment. By sourcing data from a large collection of real-world user prompts submitted to a deployed LLM API \cite{zhaowildchat}, the LENS framework pairs 1) real-world user prompts and teacher model responses as instruction tuning data with 2) real-world user prompts in plausible post-deployment shift settings as evaluation data. By using a suite of quantitative measures to estimate various aspects of distribution shift between instruction tuning and evaluation data, and by measuring model performance on shifted evaluation prompts after instruction tuning, this framework offers a systematic, controlled evaluation of model performance across multiple types and degrees of natural shifts in user prompt distributions. Using this evaluation framework, we experiment in 192 post-deployment prompt drift settings, training a total of 81 models on 4.68M training prompts, and evaluating on 57.6k total evaluation prompts, to evaluate how real-world natural prompt distribution shifts affect deployed LLM performance.

\textbf{Summary of Findings:} Using a suite of unsupervised measures of  distributional shift in natural language, we answer RQ1 by finding that natural prompt distributions shift significantly over time, between user groups, and across geographical regions (e.g. average MAUVE of 0.54 between time-delayed prompts and 0.05 between geography-separated prompts). Answering RQ2, we show that LLM performance is considerably worse in naturally-occurring OOD user prompt shift settings\footnote{We refer to natural prompt shifts as OOD settings.}. We show that natural prompt distribution shift and LLM performance degradation both increase over time (see Figure \ref{fig:result_1}), and LLM performance degradation is drastic across user groups (88\% average loss rate vs oracle models trained on OOD data, see Table \ref{tab:intermodel_results}). Our findings show that instruction following ability learned from real-world user prompts does not naturally transfer in dynamic deployment settings. These insights motivate the need for a better understanding of natural shifts in user-submitted LLM prompts (see Table \ref{tab:qualitative_examples}), and more informed methods for mitigating performance loss due to the natural evolution of user prompts.




\begin{table*}
  \centering
  \begin{tabular}{lrrrrr}\hline
    \textbf{Axis} &\textbf{OOD Eval Data} &\textbf{Oracle Data} &\textbf{ID Data} &\textbf{\# ID} &\textbf{\# Settings} \\\hline
    \textbf{Time} &Month $N$ &Months $<= N$ &Months $<= M$, $M < N$ &10 &64 \\
    \textit{Example} &\textit{Month 5} &\textit{Months 0-5} &\textit{Months 0-2}& & \\
    \hline
    \textbf{User Group} & Users $A\_i$, $V\_j$ & Users $A\_i$, $V\_j$ & $A\_k$, $V\_l$, $k \neq i$ or $l \neq j$ &8 &56 \\
    \textit{Example} &\textit{Late/High}&\textit{Late/High} &\textit{Early/Low}& & \\
    \hline
    \textbf{Geography} &Location $L\_i$ &Location $L\_i$ &Location $L\_j$, $i \neq j$ &9 &72 \\
    \textit{Example} &\textit{California} &\textit{California} &\textit{Tokyo}& & \\
    \hline
    \textbf{Total} & \textbf{19.2k Prompts} & \textbf{1.56M Prompts} & \textbf{1.56M Prompts} &\textbf{27} &\textbf{192} \\
    \hline
   \end{tabular}
   \caption{Overview of dataset splits created along each axis, resulting in 192 total realistic \{ID, OOD\} scenarios across 27 unique ID training datasets. As an example \{ID, OOD\} dataset split by user group, LENS creates ID training data from prompts collected from users who were early adopters with low volume (first 33\% of users appearing in WildChat, bottom 33\% of users by prompts per day). This is paired with OOD evaluation data collected from users who were late adopters with high volume. Oracle training data in each case contains no overlap with OOD evaluation data. Refer to Appendix \ref{app:dataset_details} for full dataset creation details.}
   \label{tab:dataset_overview}
\end{table*}


\section{Analysis Methods}
\label{sec:framework}

In this section we introduce our primary analysis method, the the \textbf{L}LM \textbf{E}valuation under \textbf{N}atural prompt distribution \textbf{S}hift (\textbf{LENS}) framework. Figure \ref{fig:framework} gives an overview of the LENS evaluation framework. First, LENS emulates real-world LLM environments by creating 192 \{ID, OOD\} (in-distribution, out-of-distribution) dataset splits sourced from real user prompts (Section \ref{sec:framework_datasets}). Next, LENS evaluates LLM performance degradation under distributional shift by comparing 81 fine-tuned ID models against fine-tuned OOD (\textit{oracle}) models for each dataset split (Section \ref{sec:framework_training}), to answer RQ2. Finally, LENS deepens understanding of LLM robustness in real-world settings by estimating the degree of natural prompt distribution shift between ID training and OOD evaluation data in each setting using natural language distribution shift measures (Section \ref{sec:framework_measures}), addressing RQ1. We examine the relationships between natural prompt distribution shift and LLM robustness across a variety of LLMs in Section \ref{sec:results}.

\subsection{Curating Prompt Shift Datasets}
\label{sec:framework_datasets}

We are interested in evaluating LLMs in the context of instruction following post-deployment, so we evaluate LLMs on actual user prompts which they might encounter post-deployment \cite{song2019using, don2025sharelm}. We do this by sourcing training and evaluation data from LLM queries collected ``in the wild.'' Our goal is to simulate realistic LLM deployment where models go through instruction fine-tuning on previously-collected user prompts and are deployed in evolving environments. As time passes, existing users may shift their behavior, and new users --- often from different demographic, geographic, or usage contexts --- begin interacting with the model. These dynamics naturally lead to changes in the prompt distribution over time, between user groups, and across regions.

We utilize WildChat \cite{zhaowildchat}, which is comprised of 1M user queries submitted to various OpenAI models, to investigate natural prompt distribution shift in a deployed LLM setting. WildChat contains real-world, natural user prompts in a deployed LLM setting. We use the user prompts and frontier model responses as instruction tuning data and evaluation data \cite{linwildbench2024}. We restrict to English prompts for simplicity, and we use the initial user message (prompt) and model response pairs. These prompts were collected by offering users free access to OpenAI models in exchange for consent to use their data. As opposed to other datasets which capture user-generated LLM queries/prompts \cite{zhenglmsys, kopf2023openassistant, don2025sharelm}, WildChat is uniquely well-suited for LENS as its data collection strategy includes IP addresses and request headers that contain timestamp metadata. Other work has similarly focused on WildChat to explore real-world LLM behavior, including user interaction trends and model response analysis \cite{brighamdeveloping, brahman2024art, rottger2025safetyprompts, zhong2024explaining, ye2024fedllm}.

 We construct 192 \{ID, OOD\} settings, including 27 unique ID datasets, using prompts and responses from WildChat in which user prompts are bucketed by time, user group, and geographical location (see Table \ref{tab:dataset_overview}). ID data is used for training each model and OOD data is used for evaluating that model in a realistic setting. Oracle training data is drawn from the same bucket as OOD evaluation data (ensuring no overlap with OOD data), and is used for training an oracle model whose responses are compared with ID model responses. Details of dataset construction can be found in Appendix \ref{app:dataset_details}\footnote{Dataset creation code can be found at \url{https://github.com/pkseeg/lens}.}.

\textbf{Over Time.} WildChat contains user queries collected from April 2023 to May 2024. As some time periods have more data than others, we sort by timestamp and split the prompts into 12 buckets which we refer to as ``months'' for simplicity. OOD evaluation datasets consist of 1k prompts sampled exclusively from each month. To simulate training on prior user queries, we construct ID training datasets by randomly sampling prompts from data up to each month. For each OOD evaluation month $N$, the corresponding oracle data is collected from up to and including month $N$, indicating training on all available data up to that point. ID data is collected on up to month $M$ where $M < N$, representing different lags in training data availability. This setup mimics a realistic deployment pipeline in which an LLM is trained on past prompts and then deployed to handle future user queries, potentially subject to natural prompt distribution shift.

\textbf{Between User Groups.} We treat anonymized IP addresses as proxies for unique users, which is a realistic setting for deployed models which often operate without persistent user accounts \cite{ye2024fedllm}. As most individual users do not submit a sufficient number of prompts to train a model or analyze performance trends, we divide users into user groups based on two latent behavioral dimensions: Adoption stage (early, medium, late), determined by when a user first appeared in the dataset; and query volume (low, medium, high), based on the rate of queries submitted per day. This results in 8 usable user groups (excluding late adopters with low volume, due to insufficient data). For each group, we sample 10k training and 1k evaluation prompts. We consider evaluation scenarios in which oracle training and OOD evaluation data come from the same user group, and ID training data comes from a different group, isolating the effect of new user groups on model performance.

\textbf{Across Locations.} WildChat uses IP-to-location mappings to assign each query a geographic region. We consider city/state-level granularity and select 9 regions with sufficient data coverage (5k for training, 1k for evaluation). These states serve as coarse representations of distinct geographical regions, e.g. California, Paris and Tokyo. We consider evaluation scenarios in which oracle training and OOD evaluation data come from the same location, and ID training data comes from a different location, allowing us to assess performance under natural geographic prompt distribution shift.

\subsection{Evaluating LLM OOD Performance}
\label{sec:framework_training}

LLM performance can be influenced by a wide variety of factors including pretraining data \cite{myntti2025register}, dataset size \cite{kaplan2020scaling}, and training techniques \cite{zhangscaling}. In studying OOD performance, we wish to isolate the effects of natural prompt distribution shift between instruction fine-tuning data and evaluation data. We do this by fixing a base model (and thus pretraining data/procedures), instruction tuning dataset size, training procedures, and model evaluation procedures. We give an overview of these processes here, and give full details in Appendix \ref{app:train_details}.

We use three pre-trained base models in our experiments. We first use Qwen2.5-7B\footnote{Qwen/Qwen2.5-7B} \cite{qwen2025qwen25technicalreport}, as this model is open-source and has been used for other work investigating instruction fine-tuning \cite{chen2024sifo, dong2025toward}. To investigate the relationship between model size and model degradation under natural prompt distribution shift, we employ the larger Qwen2.5-14B\footnote{Qwen/Qwen2.5-14B}. Finally, to compare the effects of natural prompt distribution shift on a model from a different model family, we use Llama3-8B\footnote{meta-llama/Meta-Llama-3-8B} \cite{llama3modelcard}, which has similarly been used to investigate instruction fine-tuning \cite{zheng2024llamafactory, lipreserving}. As we train 3 base models in each of 27 ID settings (see Table \ref{tab:dataset_overview}), we train 81 models total. All three models are considerably smaller than the models used to generate responses in WildChat, meaning WildChat is viable as both a learning objective, i.e. distillation of the OpenAI models in WildChat, and an evaluation setting \cite{linwildbench2024}. As we focus on measuring natural shifts in  user prompts, we choose to scale \{ID, OOD\} dataset pairs (rather than models) to get a more comprehensive view of possible natural prompt shift settings, and use three models from different sizes and families to evaluate performance under these shifts. Thus our evaluation prioritizes computational resources to answer research questions about the types and impact of shift in broad settings, rather than detailed comparisons between many different LLMs. 

In each setting, we perform supervised fine-tuning of each base model to create two fine-tuned models. For each dataset series using each base model, we train an ID model on the ID dataset and an oracle model on the oracle dataset\footnote{Training details can be found in Appendix \ref{app:train_details}.}. We directly compare the ID and oracle models by generating responses for 100 fixed prompts from the OOD evaluation data in each of the 192 splits (19.2k total for each base model). Using the same evaluation prompt as in WildChat \cite{zhaowildchat}, we compare ID and oracle model responses using GPT-4o as the judge LLM \cite{zheng2023judging}. We evaluate the performance of the ID model on OOD evaluation data by using the ID model loss rate (conversely oracle model win rate), i.e. percentage of response comparisons won by the model trained on oracle instruction tuning pairs. Thus, ID model loss rate indicates the extent to which sourcing training data from the OOD setting is better than ID (i.e. excluding ties). For example, a Qwen2.5-7B model trained on data from months 0-3 loses 53\% of head-to-head comparisons on OOD evaluation data in month 7 against an oracle model trained on data from months 0-7, and ties 15\%, hence the model is winning only 30\% of these comparisons and is worse than the oracle model. The greater the ID model loss rate, the greater the disparity between the ID and oracle models, thus the less performant the ID data in deployment. We follow other works in using LLM performance as the evaluation metric for training data quality \cite{sun2024evaluating, zhuang2025meta, chen2023maybe}. 

\subsection{Measures of Natural Prompt Distribution Shift}
\label{sec:framework_measures}

Among large-scale robustness evaluations of LLMs \cite{yuan2023revisiting, wenzel2022assaying}, a critical missing element is the degree to which the OOD setting differs from the training setting. Not all distribution shifts are created equal, and faulty assumptions of distributional shift may lead to misleading estimates of model robustness. However, measuring distributional shift in natural language presents unique challenges due to the lack of a known ground-truth distribution over text inputs \cite{lebrunevaluating}. Unlike structured data with explicit feature spaces, natural language is inherently high-dimensional, sparse, and context-dependent. As a result, quantifying natural prompt distribution shift requires proxy measures that capture meaningful variation between text corpora. Prior work has leveraged supervised approaches to detect shift in known prompt properties such as complexity \cite{he2024guiding}, toxicity \cite{zhenglmsys}, or topic \cite{myntti2025register}. However, these methods can be impractical for deployed LLMs, where future prompts may diverge in unforeseen and unconstrained ways \cite{jang2024driftwatch}. To address this challenge, the LENS framework emphasizes unsupervised measures of distributional shift in natural language. We briefly describe four measures used in our analysis here, describing full details and additional measures in Appendix \ref{app:measure_details}.

\begin{figure*}
    \centering
    \includegraphics[width=\textwidth]{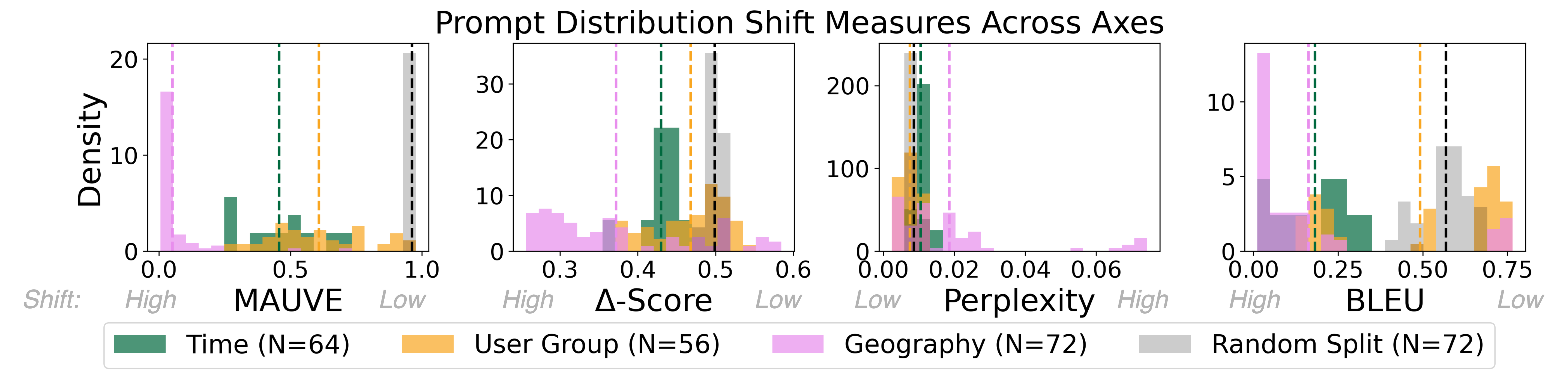}
    \caption{We estimate natural prompt distribution shift between all \{ID, OOD\} settings using four distribution shift measures, finding that prompt distributions shift significantly over time, between user groups, and across geographical locations. We also find that MAUVE can best distinguish between distributions, as it has a high disparity between estimates across axes and the random baseline estimates, which are more precise and stable.}
    \label{fig:shift}
\end{figure*}

\textbf{Notation} Let $P$ denote the distribution of pre-deployment training user prompts, and $Q$ denote the distribution of post-deployment user prompts. We call $X = \{X_i\}_{i=1}^n \sim P$ the sample of $n$ training prompts, and $Y = \{Y_i\}_{i=1}^m \sim Q$ the sample of $m$ post-deployment prompts. To measure shift between training prompts and post-deployment prompts, we will use sample-based divergence estimators $D(P,Q)$. In practice, we use $n=m=1000$ samples from pre- and post-deployment prompt pools \cite{seegmiller2023statistical}. The $n=1000$ pre-deployment training prompts are taken as a subset from the ID training data and the $m=1000$ post-deployment prompts are the OOD data described in Section \ref{sec:framework_datasets}.

\textbf{MAUVE}, introduced by \citet{pillutla2021mauve}, is an embeddings-based measure designed for text generation evaluation                    , which compares a pair of text distributions by computing information divergences in a quantized embedding space. MAUVE simultaneously captures the probability of Type I and Type II errors in language generation --- thus repurposing MAUVE to measure divergence between prompt distributions $D(P,Q)$ is meant to estimate the overlap in relative frequency of prompt characteristics in $P$ and $Q$. A MAUVE score closer to 0 indicates higher shift between prompt distributions.

$\Delta$\textbf{-score}, introduced by \citet{seegmiller2023statistical},\footnote{The original authors use $Q$-score as notation, but we use $\Delta$-score for notational clarity in this work.} is an embeddings-based measure which uses a similarity-based statistical depth $D_\delta(x_i, P)$ to estimate how representative a text $x_i \sim P$ is of the distribution. The goal of the $\Delta$-score is to capture how well one text distribution represents another. A $\Delta$-score of 0.5 indicates an equal likelihood of representation, and a lower $\Delta$-score indicates higher shift.

\textbf{Perplexity} is a language model-based measure that quantifies how well a probability distribution predicts a sample. Lower perplexity values indicate that the model assigns higher probabilities to the observed sequences, suggesting better predictive performance. The LENS framework utilizes the LLM itself, fine-tuned on samples $\{X_i\}_{i=1}^n \sim P$ to estimate perplexity of samples $\{Y_i\}_{i=1}^n \sim Q$. This approach captures how \textit{surprising} prompts from the $Q$ distribution are based on $P$.

\textbf{BLEU}, originally developed for machine translation evaluation \cite{papineni-etal-2002-bleu}, provides a statistical measure of $n$-gram overlap between text distributions. Higher BLEU scores indicate greater lexical similarity between distributions, capturing surface-level textual patterns rather than deep semantic relationships. BLEU offers an estimate of lexical divergence between prompt distributions, measuring explicit differences in vocabulary usage, phrase construction, and local contextual patterns.

\section{Evaluation and Results}
\label{sec:results}

In this section we quantify the magnitude of natural prompt distribution shift in Section \ref{sec:results_diverge} (RQ1), then we examine the degrading impact of this shift on LLM performance in Sections \ref{sec:results_degrade} and \ref{sec:results_degradation_comp} (RQ2). Results in Sections \ref{sec:results_diverge} and \ref{sec:results_degrade} are averaged across three models (Llama3-8B, Qwen2.5-7B, and Qwen2.5-14B) to ensure robustness; Section \ref{sec:results_degradation_comp} provides a detailed inter-model comparison.

\begin{figure*}
    \centering
    \includegraphics[width=\textwidth]{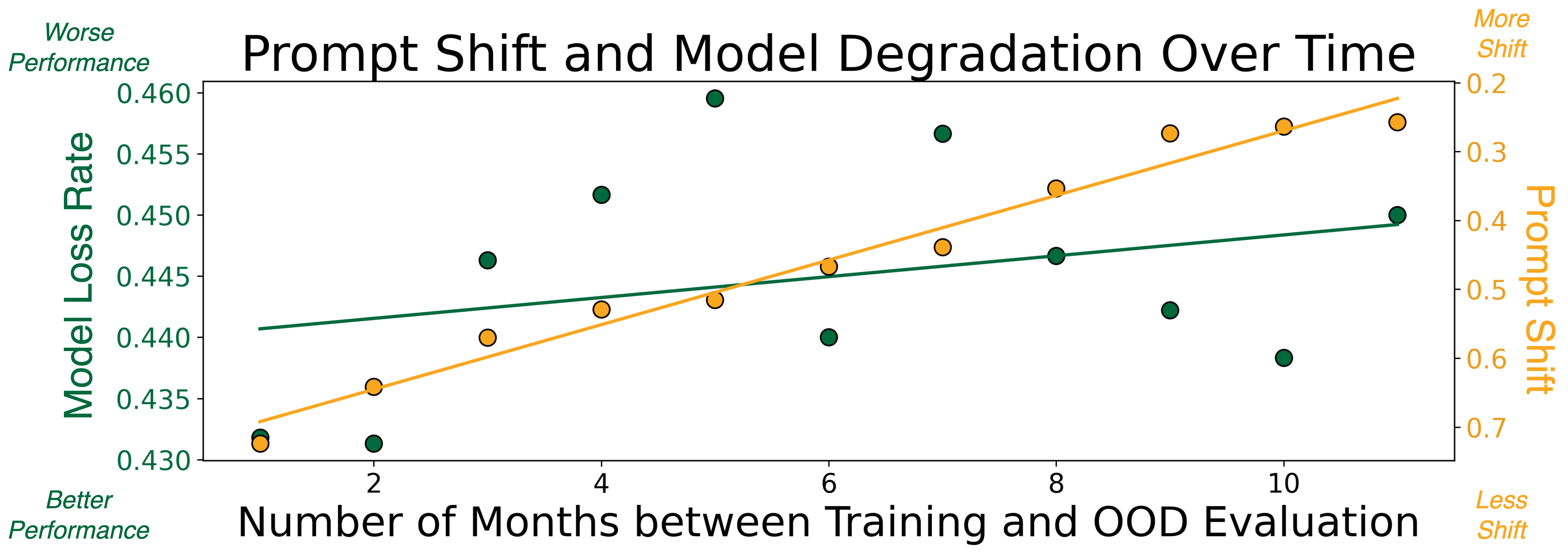}
    \caption{Using the LENS framework to evaluate LLM performance in \textbf{64} time-shifted deployment settings, averaging across 3 different LLMs, we find that significant distributional shift (measured in MAUVE) occurs in user prompts submitted to a deployed LLM over time. We find that the degree of \textbf{natural prompt distribution shift} increases as the lag between training data and time-delayed OOD evaluation data increases. We also find that LLMs trained on user prompts perform poorly  on time-delayed OOD evaluation data (measured in model loss rate vs an oracle model trained on OOD data), and that the performance loss tends to be more drastic as the lag increases.  See Section \ref{sec:results_degrade} for more discussion of these results.}
    \label{fig:result_1}
\end{figure*}

\subsection{Distribution Shift in Real-World LLM Queries}
\label{sec:results_diverge}

The LENS framework measures natural prompt distribution shift using the unsupervised measures described in Section \ref{sec:framework_measures}: MAUVE, $\Delta$-score, perplexity, and BLEU. Shifts are estimated between training and OOD evaluation data in each of the 192 \{ID, OOD\} natural settings described in Section \ref{sec:framework_datasets} along three axes: time, user group, and geographical location. To compare these estimates against a baseline amount of shift which would be expected in random sampling, we also measure shift between $72$ dataset pairs randomly sampled from all English WildChat prompts.

As shown in Figure \ref{fig:shift}, significant natural prompt distribution shifts are observed across all three axes. Notably, the geographical location axis exhibits the strongest shift, indicating that different geographical behavioral patterns are strongly associated with changes in prompt characteristics. Natural prompt distribution shifts are also strong between time-shifted datasets, e.g. with an average MAUVE score of 0.54 vs the random baseline MAUVE score of 0.96. While user groups based on adoption rate and usage volume exhibit the least natural prompt distribution shift among the three axes which we evaluate, the average MAUVE (0.61), $\Delta$-score (0.47), and BLEU (0.49) scores in this group still indicate measurable distributional shift compared to baseline scores (0.96, 0.50, 0.57 average MAVUE, $\Delta$-score, and BLEU, respectively). These findings highlight that prompt distributions encountered by deployed LLMs can vary substantially depending on when the model is used and who is using it. We find that perplexity estimates shift between datasets to be closer to the random baseline. We hypothesize that perplexity measurements are impacted by the vast amounts of pre-training data seen by the LLM, impacting the surprisal of the model on OOD data. However, as we see in Table \ref{tab:intermodel_results}, pre-training does not mitigate performance degradation in the context of natural prompt distribution shift. Shift estimates can be found in tabular format in Table \ref{app:tab:shift} in Appendix \ref{app:shift_results}.


\subsection{LLM Performance Degradation Due to Natural Prompt Distribution Shift}
\label{sec:results_degrade}

The LENS framework evaluates the relationship between natural prompt distribution shift and LLM performance by measuring LLM OOD performance in realistic post-deployment settings. For each prompt in the OOD evaluation dataset, we collect responses from two models: the ID model (i.e., the model trained on data representative of expected pre-deployment user prompts), and the oracle model trained on data sourced from the same bucket as the OOD evaluation dataset. Following \citet{zhaowildchat}, we use GPT-4o to judge the responses. We measure ID model loss rate compared to the oracle model to estimate LLM OOD performance degradation.

\begin{table*}
\centering
\begin{tabular}{lrcccccc}\hline
\textbf{Axis} & \textbf{Model} & \textbf{MAUVE} & \textbf{$\Delta$-score} & \makecell{\textbf{Loss Rate} \\ \textbf{vs Oracle}} & \makecell{\textbf{Win Rate} \\ \textbf{vs Oracle}} & \makecell{\textbf{Tie Rate} \\ \textbf{vs Oracle}} \\
\hline
Time &L8B &0.5125 (0.18) &0.4963 (0.02) &0.4388 (0.05) &0.3991 (0.05) &0.1359 (0.03) \\
Time &Q7B &0.5437 (0.18) &0.4175 (0.04) &0.4409 (0.05) &0.3553 (0.04) &0.1770 (0.04) \\
Time &Q14B &0.5510 (0.17) &0.4049 (0.05) &0.4509 (0.05) &0.3573 (0.05) &0.1630 (0.04) \\
\hline
UG &L8B &0.5906 (0.20) &0.4666 (0.03) &0.8466 (0.04) &0.0196 (0.02) &0.1166 (0.03) \\
UG &Q7B &0.6139 (0.19) &0.4689 (0.07) &0.8780 (0.02) &0.0173 (0.01) &0.0880 (0.03) \\
UG &Q14B &0.6217 (0.18) &0.4685 (0.08) &0.8923 (0.03) &0.0170 (0.01) &0.0730 (0.02) \\
\hline
Geo &L8B &0.0471 (0.10) &0.3642 (0.09) &0.8778 (0.05) &0.0099 (0.01) &0.0917 (0.04) \\
Geo &Q7B &0.0529 (0.11) &0.3787 (0.11) &0.8849 (0.04) &0.0104 (0.01) &0.0864 (0.04) \\
Geo &Q14B &0.0546 (0.11) &0.3737 (0.10) &0.8833 (0.04) &0.0107 (0.01) &0.0871 (0.05) \\
\hline
\end{tabular}
\caption{MAUVE, $\Delta$-score, and performance against oracle model across models and prompt shift axes. MAUVE (0-1) measures distributional similarity between ID and OOD prompt distributions by estimating Type I and Type II errors in language generation (higher is more similar). $\Delta$-score (0-0.5) measures the likelihood that a random OOD prompt is representative of the ID distribution (0.5 implies no shift, lower is more shift). Loss/Win/Tie Rate reflect the proportion of head-to-head response comparisons in which the model trained on shifted-distribution prompts performs worse than, better than, or equivalently to the oracle model (trained on prompts from the same distribution as the evaluation data). Models: Llama3-8B (L8B), Qwen2.5-7B (Q7B), Qwen2.5-14B (Q14B). Axes: Time, User Group (UG), and Geography (Geo). All three natural shift axes degrade performance, with an average loss rate of 73\%, and this degradation is consistent across model architecture and size. Scores are mean (standard deviation).}
\label{tab:intermodel_results}
\end{table*}

Figure \ref{fig:result_1} presents results along the time axis, averaged by the number of months between ID training and OOD evaluation settings. We see a correlation between the magnitude of natural prompt distribution shift and the degradation in LLM performance, both of which increase over time. We see a similar pattern in user group and geography settings, which we detail in Appendix \ref{app:ug_geo_results}. Aggregated performance metrics can be found in tabular format in Table \ref{app:tab:performance} in Appendix \ref{app:performance_results}.


\subsection{Inter-Model Performance Degradation}
\label{sec:results_degradation_comp}

Table \ref{tab:intermodel_results} gives distribution shift (MAUVE and $\Delta$-score\footnote{Perplexity and BLEU scores are given in Appendix \ref{app:shift_results}.}) and performance (loss, win, and tie rates vs oracle model as described in Section \ref{sec:framework_training}) results for each model, averaged for each natural prompt shift axis. We see that the prompt shifts found in Section \ref{sec:results_diverge} are consistent across model families and sizes. With average loss rates of 0.44, 0.88 and 0.88 compared to win rates of 0.37, 0.02, and 0.01 for time, user group, and geographical shift settings, respectively, performance degradation in OOD settings is consistent in all three models, regardless of model size or family. We find that performance drop across user group and geographical splits is particularly high, with average loss rates of 87\% and 88\%, indicating that instruction following ability generalizes poorly across user groups and geographical splits. These results indicate that instruction following ability learned from real-world user prompts does not transfer in natural prompt shift settings.

\subsection{Towards Identifying Underlying Prompt Shift Trends}
\label{sec:results_trends}

\begin{table*}
\centering
\small
\begin{tabular}{m{1.5cm} m{6.5cm} m{6.5cm}}
\hline
\textbf{Axis} & \textbf{ID Prompt} & \textbf{OOD Prompt} \\
\hline
Time        & (Month 4) Make up sentences with this word: impartial & (Month 9) As a prompt generator... you will create image prompts for the AI to visualize... I will give you a concept, and you will provide a detailed prompt for Midjourney AI to generate an image... Please adhere to the structure and formatting below, and follow these guidelines...\\
\hline
User Group  & (High Volume) give me a response to \{message\} to send in a discussion, VERY SHORT, CONCISE \& CLEAR. ONLY RETURN THE RAW MESSAGE, DO NOT SAY "Hey here is the message you asked" & (Medium Volume) write some Eating disorders that will be taught for nutritional biochemistry course \\
\hline
Geography  & (Massachusetts) write a comedic and vividly detailed story set in the TV show Z Nation about 10K  ... & (Paris) explique de manière claire pour ma documentation ma pipeline ... \\
\hline
\end{tabular}
\caption{Representative ID/OOD prompt pairs illustrating qualitative shift trends along each natural prompt distribution shift axis. We find that prompts instructions tend to become more specific over time, that high volume usage is consistent with more templated prompts, and that geographically distinct prompts contain cultural and linguistic differences.}
\label{tab:qualitative_examples}
\end{table*}

We have shown that instruction following ability learned from real-world user prompts does not naturally transfer in dynamic deployment settings. To motivate future investigation into the underlying drivers of this natural prompt shift, we conduct an initial qualitative analysis of 30 representative\footnote{Details of the sample selection and labeling process is given in Appendix \ref{app:prompt_shifts}, along with three examples per axis.} prompt pairs sampled from ID and OOD splits along each axis (90 total). Representative examples from each axis are shown in Table \ref{tab:qualitative_examples}.

\textbf{Over Time.} Representative prompts from later time periods tended to be longer and more structurally elaborate, with explicit formatting constraints and instructions, while earlier prompts were more exploratory and casual. These observations are consistent with prior findings that user behavior evolves as familiarity with model capabilities grows~\cite{ma2024death, tafreshipour2025prompting}, suggesting that temporal prompt shift may reflect changing user expectations rather than purely topical drift.

\textbf{Between User Groups.} Representative prompts across user groups revealed systematic differences in how distinct populations conceptualize and engage with the model. High-volume users, for example, showed evidence of templated, repetitive prompting, suggesting workflow integration rather than exploratory use---for example, variations of the templated ID prompt in row 2 of Table \ref{tab:qualitative_examples} was sampled several times in our manual analysis. Low-volume users exhibited more eclectic, less patterned prompt distributions spanning a wider range of domains and task types. These patterns suggest that user group-level shift may reflect systematic differences in how distinct user populations conceptualize and engage with deployed LLMs, consistent with the high ID model loss rates (avg.\ 88\%) observed for this axis in Section~\ref{sec:results_degradation_comp}.

\textbf{Across Geographic Regions.} In our manual analysis, representative prompts from different geographic regions tended to reflect various cultural differences (see example in Table \ref{tab:qualitative_examples}), in addition to linguistic variations potentially stemming from the use of English as a second language or code-mixing \cite{yang2025codemixbench}.

Our findings motivate future work on systematic taxonomies of prompt shift categories and supervised probes for interpretable shift decomposition. More broadly, our analysis suggests that evaluating and mitigating natural prompt distribution shift is a general challenge for any deployed LLM operating over an evolving user base. Robust evaluation under natural prompt shift will become increasingly important for LLM reliability as deployment continues to scale and users continue to evolve.
\section{Related Works}

\textbf{OOD Robustness in LLMs} Recent work has argued that many OOD tests do not represent a sufficient probing of LLM robustness \cite{Gupta2023WhispersOD, teney2023id}. This is a pronounced issue in NLP, where the definition of OOD data can be less clear-cut compared to other machine learning settings, partially because of the difficulty of defining a natural distributional model of natural language \cite{lebrunevaluating, ilia2024predict}. Some prior works have presented thorough evaluations of OOD robustness in LLMs by systematically selecting ID and OOD dataset pairs across a wide variety of common tasks, including some in NLP \cite{yuan2023revisiting, wenzel2022assaying}. These focus on task-specific OOD robustness, where the LENS framework focuses on OOD robustness to user prompting in deployment and measures distribution shift in an unsupervised manner.

\textbf{Data-Centric Distribution Shift Evaluation in NLP} Prior works have investigated temporal shift in language processing tasks such as document classification, named entity recognition, language modeling, and sentiment classification \cite{agarwal2022temporal, biesialska2020continual, loureiro2022timelms, lazaridou2021mind, huang2018examining, jang2024driftwatch, luu2022time}. Other works have analyzed LLM performance by assessing the relationships between pre-training data term frequencies and downstream performance \cite{mccoy2023embers, razeghi2022impact, pmlr-v202-kandpal23a, elazar2022measuring}. Other works have investigated robustness of LLMs to changes in instructions \cite{sun2024evaluating, mizrahi2024state, gu2023robustness}. We are the first to measure natural prompt shifts and examine the impacts on LLMs in deployment.

\section{Conclusion}
We introduced the \textbf{L}LM \textbf{E}valuation under \textbf{N}atural prompt \textbf{S}hift (\textbf{LENS}) framework for analyzing the relationship between instruction tuning dataset distributions and user prompt distributions in deployment. We find that natural prompt distribution shifts correlate with LLM post-deployment performance degradation. We provide the LENS framework and our findings to facilitate future research in evaluating and mitigating LLM performance loss under natural prompt distribution shift.

\section{Limitations}
Large scale evaluation of LLMs is expensive: our evaluation required an estimated 1,200 GPU hours on a single A100 between model training, response generation, and model-based shift measurement. We choose to scale shift settings and models, but this limits our investigation into explicit mixing of distributions for diversity \cite{yuan2023revisiting}, and intersectionality (e.g. specific user groups over time), which may be of interest in specific domains \cite{devinney2024we}. We are limited by the availability of real-world user prompt datasets with temporal and user designations \cite{zhaowildchat, zhenglmsys, kopf2023openassistant}. We follow WildChat \cite{zhaowildchat} in using LLM-As-Judge in our evaluation, which is known to have some limitations \cite{zheng2023judging, krumdick2025no}.




\section*{Acknowledgments}
This work was supported in part by NIH grant 1R21DA059665-01A1, Clinical and Translational Science Institute grant 1UM1TR004772, a Google Research Scholar award, and the NSF Research Traineeship, Transformative Research, and Graduate Education in Sensor Science, Technology and Innovation grant (DGE-2125733).

\bibliography{custom}

\appendix

\newpage

\appendix

\section{Complete Dataset Collection Details}
\label{app:dataset_details}

\begin{figure*}
    \centering
    \includegraphics[width=\textwidth]{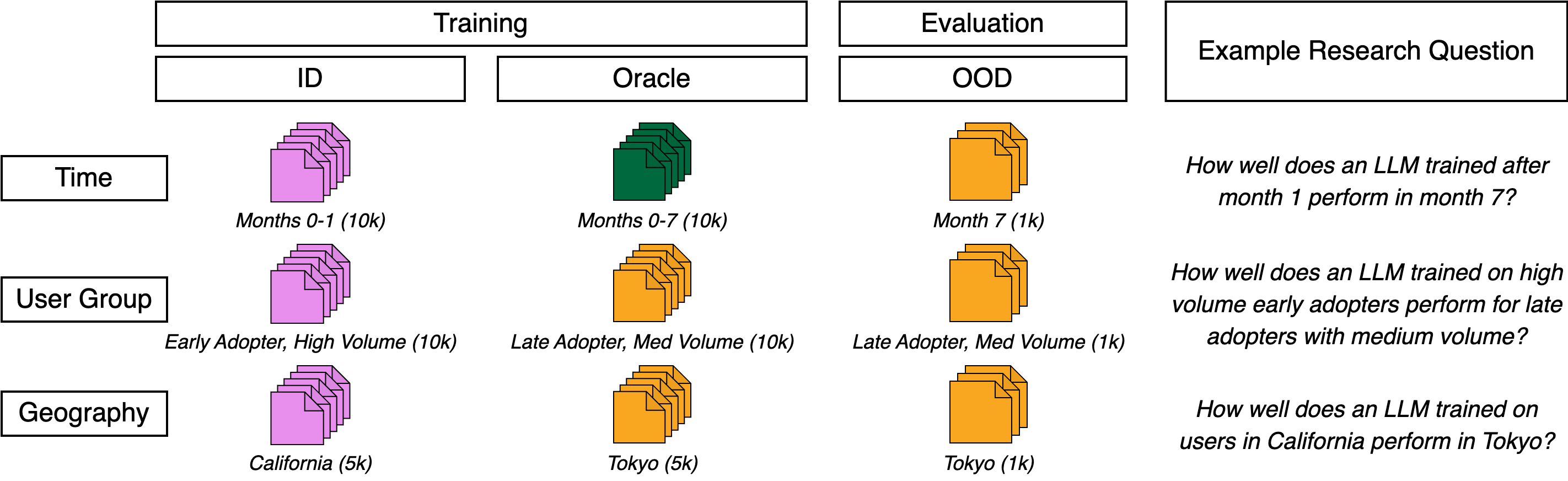}
    \caption{We bucket the WildChat \cite{zhaowildchat} dataset across three axes to create 192 natural prompt shift settings, training a total of 81 models on 4.68M training instances, and evaluating on 57.6k total evaluation instances. Here we give examples of each natural prompt shift setting, along with motivating research questions for each.}
    \label{fig:app:dataset}
\end{figure*}

To prepare the WildChat dataset for use in our experiments, we begin with the version which has had toxic conversations, and messages containing PII, removed from the dataset\footnote{\url{https://huggingface.co/datasets/allenai/WildChat-1M}}. We subset to obtain only the 479k conversations which were classified to be in the English language, and we also take only unique prompts. To ensure that we have \{query, response\} pairs available for LLM instruction fine-tuning, we subset to consider only the first user prompt and the first response from the OpenAI model. This leaves us with a dataset containing 479k \{prompt, model response\} pairs which are each associated with the timestamp of the initial message and the hashed IP address of the user. As seen in Figure \ref{fig:app:dataset}, we then split the dataset across three axes. 

\textbf{Time} To bucket the dataset across time, we sort the dataset by timestamp and divide it into 12 equally-sized buckets, which we refer to as ``months'' for simplicity. To create \{ID, OOD\} pairs, we are interested in evaluating how well an ID training dataset, comprised of all data up to a given timestamp, performs in an OOD setting defined by a certain ``lag.'' We randomly sample 1k OOD evaluation data from each month $N \in \{1, \dots, 11\}$. Oracle data for each OOD evaluation setting is a randomly sampled 10k rows from all months up to and including $N$, i.e. $0 <= Orcl <= N$, ensuring that there is no overlap between OOD eval and oracle training data. This represents the no-lag scenario. ID data is then randomly sampled from each possible collection of months $0 <= M$, where $M < N$. This is the scenario in which some amount of lag is introduced, and enables us to investigate the difference in performance between models trained on $lag = N - M$ data (ID) and no lag data (oracle).

\textbf{User Group} One reason that prompt distributions may change in deployment is the introduction of new users. We treat anonymized IP addresses as proxies for unique users \cite{ye2024fedllm}. As most individual users do not submit a sufficient number of prompts to train a model or analyze performance trends in isolation, we divide users into user groups based on two latent behavioral dimensions: Adoption stage (early, medium, late), determined by when a user first appeared in the dataset; and Query volume (low, medium, high), based on the rate of queries submitted per day. The three buckets for adoption rate and query volume are defined by the 33rd and 66th percentiles.  This results in 8 usable user groups (excluding late adopters with low volume, due to insufficient data). For each group, we sample 10k training and 1k evaluation prompts. We consider evaluation scenarios in which oracle training and OOD evaluation data come from the same user group, and ID training data comes from a different group, allowing us to isolate the effect of introducing new user groups on model performance.

\textbf{Geography} WildChat uses IP-to-location mappings to assign each query a geographic region. These regions are recognized at the national and state/city level. We consider city/state-level granularity and select 9 regions with sufficient data coverage (5k for training, 1k for evaluation). These states serve as coarse representations of distinct geographical regions and consist of California, Krasnodar Krai, Massachusetts, Michigan, Moscow, New York, Paris, Pennsylvania, and Tokyo. We consider evaluation scenarios in which oracle training and OOD evaluation data come from the same location, and ID training data comes from a different location, allowing us to assess performance under natural geographic prompt distribution shift.

\section{Instruction Tuning Details}
\label{app:train_details}


As highlighted in Section \ref{sec:framework_training}, we use three pre-trained base models in our experiments. We first use Qwen2.5-7B\footnote{Qwen/Qwen2.5-7B} \cite{qwen2025qwen25technicalreport}, as this model is open-source and has been used for other work investigating instruction fine-tuning \cite{chen2024sifo, dong2025toward}. To investigate the relationship between model size and model degradation under natural prompt distribution shift, we employ the larger Qwen2.5-14B\footnote{Qwen/Qwen2.5-14B}. Finally, to compare the effects of natural prompt distribution shift on a model from a different model family, we use Llama3-8B\footnote{meta-llama/Meta-Llama-3-8B} \cite{llama3modelcard}, which has similarly been used to investigate instruction fine-tuning \cite{zheng2024llamafactory, lipreserving}. By fixing a base model in our evaluation, we can consider pretraining data and procedures to be fixed, enabling us to isolate the effects of natural prompt distribution shift on downstream performance. 

In each case where a model is trained, we utilize LLaMA-Factory \cite{zheng2024llamafactory}\footnote{https://github.com/hiyouga/LLaMA-Factory} to standardize our training procedure. We train for 1 epoch using a batch size of $4$ on a single A100 (80GB) GPU. We estimate each training run requires an average of $2$ GPU hours, for a total of 774 GPU hours for model training.  

We train using low-rank adaptation (LoRA) \cite{hu2022lora}, which has shown to be a performant and efficient fine-tuning strategy \cite{shulman2025lora, zhao2024lora}. We configure our LoRA adaptation with a rank of $8$ and an alpha scaling factor of $16$. We use the AdamW optimizer with weight decay of $0.01$, linear learning rate scheduler with warmup over 10\% of the training steps, and gradient clipping at a norm of $1.0$. We apply mixed precision training using float16 to optimize memory usage and training speed.

\section{Detailed Discussion of Distributional Shift Measures in Natural Language}
\label{app:measure_details}

We detail three broad categories of distribution shift measures for natural language, including the four used in our analysis in Section \ref{sec:results} as well as several others which could be used within the LENS framework to analyze different types of natural prompt distribution shift. Many of these measures are designed for purposes other than distributional shift estimation, for example evaluation of open-ended text generation. However, all of these metrics operate only on paired natural language distributions $P$ and $Q$, thus they may be repurposed to measure the shift between \textit{any} two natural language distributions.

\textbf{Embeddings-based measures} utilize semantically-meaningful multidimensional text representations combined with traditional measures of divergence to measure shifts in language distributions. In order to not rely on the bias of an external embedding model, the LENS framework utilizes the LLM itself to embed user prompts by employing mean pooling of of the hidden state activations from the final layer of the model's encoder \cite{wang2024improving}. This approach captures the model's own semantic representation of the text, aligned with the LLM's internal language processing mechanisms rather than introducing potentially misaligned representations from separate embedding models.  

Pillutla et al. \cite{pillutla2021mauve} introduce embeddings-based measure \textbf{MAUVE} for measuring open-ended text generation, which compares a pair of text distributions by computing information divergences in a quantized embedding space. For a series of mixture distributions $R_\lambda = \lambda P + (1-\lambda)Q$, MAUVE estimates the area under the divergence curve (created by varying $\lambda \in (0,1)$) by quantizing the embedding space of embedding samples for efficient KL divergence calculation. The goal of MAUVE is to simultaneously capture the probability of Type I and Type II errors in machine language generation --- thus repurposing MAUVE to measure divergence between prompt distributions $D(P,Q)$ is meant to estimate the overlap in relative frequency of prompts in $P$ and $Q$. A MAUVE score closer to 0 indicates higher shift between prompt distributions.

Seegmiller et al. \cite{seegmiller2023statistical} introduce the $\Delta$\textbf{-score}, which uses a similarity-based statistical depth $D_\delta(x_i, P)$ to estimate how representative a text $x_i \sim P$ is of the distribution. The $\Delta$-score, $\Delta(P, Q)$, estimates the probability that a randomly selected text $y_i$ from $Q$ will be more representative of distribution $P$ than a randomly-selected text $x_i$ from $P$, i.e. $\Delta(P, Q) = Pr[D_\delta(X, P) \leq  D_\delta(Y, P)]$. The goal of the $\Delta$-score is to capture how well one text distribution represents another. A $\Delta$-score of 0.5 indicates an equal likelihood of representation, and a lower $\Delta$-score indicates higher shift between prompt distributions.

Other embeddings-based measures which may be used within the LENS framework include \textbf{EigenDivergence} \cite{abdulaalbalancing}, which was originally designed for detecting hallucinations by measuring semantic consistency across LLM outputs, and \textbf{average minimum distance} \cite{binu2022multi}, which estimates the smallest distance between two sets of text embeddings.

\textbf{Language model measures} utilize a reference language model to estimate token sequence likelihoods, then estimate shift by utilizing token sequence likelihoods.

\textbf{Perplexity} is a fundamental measure in language modeling that quantifies how well a probability distribution predicts a sample. Perplexity evaluates a trained model's ability to predict text sequences by computing the exponential of the average negative log-likelihood of tokens. Formally, perplexity is defined as $PPL_{P}(Q) = \exp\left(-\frac{1}{n} \sum_{i=1}^{n} \log p_{P}(y_i | y_{<i})\right)$. Lower perplexity values indicate that the model assigns higher probabilities to the observed sequences, suggesting better predictive performance. Again, in order to not rely on the bias of an external reference model, the LENS framework utilizes the LLM itself, fine-tuned on samples $\{X_i\}_{i=1}^n \sim P$ to estimate perplexity of samples $\{Y_i\}_{i=1}^n \sim Q$. This approach captures how surprising prompts from the $Q$ distribution appear from the perspective of $P$. 

\begin{table*}
  \caption{We estimate natural prompt distribution shift between all 192 \{ID, OOD\} settings along with 72 random sampling settings, using four distribution shift measures averaged across three models. We find that prompt distributions shift significantly above average over time, between user groups, and across geographical location. Results are reported as mean (standard deviation). Results for each of the 192 individual splits can be found in csv format in supplementary materials.}
  \label{app:tab:shift}
  \centering
    \begin{tabular}{lrrrrr}\hline
    \textbf{Axis} &\textbf{MAUVE} &\textbf{$\Delta$-Score} &\textbf{Perplexity} &\textbf{BLEU} \\\hline
    Time &0.5358 (0.17) &0.4396 (0.05) &0.0100 (0.01) &0.2352 (0.18) \\
    User Group &0.6088 (0.19) &0.4680 (0.06) &0.0075 (0.01) &0.4923 (0.24) \\
    Geography &0.0516 (0.11) &0.3722 (0.10) &0.0186 (0.02) &0.1627 (0.24) \\
    Random &0.9618 (0.01) &0.4991 (0.01) &0.0086 (0.01) &0.5689 (0.07) \\\hline
    \end{tabular}
\end{table*}

Other language model measures which may be used within the LENS framework include \textbf{zero-shot concatenation perplexity} \cite{lee2024crafting}, which prompts a pre-trained language model with concatenated samples from each distribution and estimates the pre-trained model's expectation of one sample following the other; \textbf{KL divergence} and \textbf{reverse KL divergence} \cite{liu2024online, bu2016estimation}, which measure the information loss when one probability distribution is used to approximate another, with KL divergence penalizing the model for placing mass where the reference distribution has little mass, and reverse KL divergence penalizing the model for failing to place mass where the reference distribution has significant mass; and \textbf{JSD divergence} \cite{lu2020diverging} which provides a symmetric alternative to KL divergence by measuring the average distance from each distribution to their mixture, making it particularly useful for comparing distributions that have minimal overlap.

\textbf{Statistical measures} estimate divergence between prompt distributions based on explicit token-level patterns and frequency distributions. 

\textbf{BLEU}, originally developed for machine translation evaluation \cite{papineni-etal-2002-bleu}, provides a statistical measure of $n$-gram overlap between text distributions. While traditionally used to compare candidate translations against reference translations, we repurpose BLEU to quantify the similarity between prompt distributions $P$ and $Q$ by calculating the geometric mean of modified $n$-gram precision scores across different $n$-gram lengths. Specifically, BLEU computes the proportion of $n$-grams in $X$ that appear in $Y$, with a brevity penalty applied to account for length discrepancies. Higher BLEU scores indicate greater lexical similarity between distributions, capturing surface-level textual patterns rather than deep semantic relationships. This approach offers complementary insights to embedding-based and reference-model-based measures and is particularly sensitive to structural and phrasal overlap. By examining token overlap at various $n$-gram levels ($n=1$ to $4$), BLEU offers an estimate of lexical divergence between prompt distributions, measuring explicit differences in vocabulary usage, phrase construction, and local contextual patterns.

A common approach to estimating training dataset/evaluation dataset overlap in NLP tasks is to use \textbf{n-gram overlap} to determine whether evaluation samples exist in the training data \cite{mccoy2023embers, razeghi2022impact, pmlr-v202-kandpal23a, elazar2022measuring}. This approach is scalable, and provides a straightforward method for detecting exact or near-exact duplicates, though it may fail to identify semantic duplicates or more complex forms of information leakage that don't preserve the same surface-level text patterns. 

Other statistical measures include token rank frequency (\textbf{Zipf}) \cite{holtzmancurious}, which takes the KL divergence between discrete token rank frequency distributions, a well-known property of human text, and \textbf{word mover score} \cite{zhao2019moverscore}, which computes the minimum cost of transforming a text from one distribution to a text in another distribution.

\section{Towards Identifying Underlying Prompt Shift Trends}
\label{app:prompt_shifts}

\begin{table*}
    \centering
    \small
    \begin{tabular}{m{1.5cm} m{6.5cm} m{6.5cm}}
    \hline
    \textbf{Axis} & \textbf{ID Prompt} & \textbf{OOD Prompt} \\
    \hline
    Time & (Month 4) Make up sentences with this word: impartial
         & (Month 9) As a prompt generator for a generative AI called ``Midjourney''... I will give you a concept, and you will provide a detailed prompt ... \\
    \hline
    Time & (Month 3) Write short 10 tweets: look at this beach right now
         & (Month 5) [Format your response using markdown. Use headings, subheadings, bullet points, and bold to organize the information]
If ``Mom 2'' is when the first mother dies and the father remarries ... \\
    \hline
    Time & (Month 3) Describe how DNA methylations can be passed during replication.
         & (Month 5) Write an article on ``Your Partner in Wellness'' ... Make sure that you don't follow ai pattern .. it should be as written with an intermediate level writer ... Also add bold and italic.
 \\
    \hline
    User Group & (High Volume) give me a response to \{message\} to send in a discussion, VERY SHORT, CONCISE \& CLEAR. ONLY RETURN THE RAW MESSAGE, DO NOT SAY ``Hey here is the message you asked''
               & (Medium Volume) write some eating disorders that will be taught for a nutritional biochemistry course \\
    \hline
    User Group & (High Volume) give me a response to \{message\} to send in a discussion, VERY SHORT, CONCISE \& CLEAR. ONLY RETURN THE RAW MESSAGE, DO NOT SAY ``Hey here is the message you asked''
               & (Medium Volume) I don't suggest, your words is under arest! ha-ha. \\
    \hline
    User Group & (High Volume) MS SQL --- Procedural Integrity Constraints, Declarative Integrity Constraints, Not Null, Unique, Default and Check constraints, Primary Key and Referential Integrity ...
               & (Medium Volume) Based on this job description for a PhD position ... write a CV for my past 3 years of work experience with 3--5 bullet points per role ... \\
    
    \hline
    Geography   & (Pennsylvania) write a script about bobby petrino losing to north dakota state 45-28 
                & (Massachussetts) Make a vividly detailed story taking place in Ancient Rome about a Roman Emperor... \\
    \hline
    Geography  & (Massachusetts) write a comedic and vividly detailed story set in the TV show Z Nation about 10K, before he met Warren's group ...
               & (Paris) explique de manière claire pour ma documentation ma pipeline ... \\
    \hline
    Geography  & (Pennsylvania) write a script about georgia southern college asun
               & (Moscow) Alice Flamand is young married socialite in 1973 who is attacked by unknown assailants ... \\
    \hline
    \end{tabular}
    \caption{Additional examples of representative prompts pairs along natural prompt distribution shift axes.}
\label{tab:qualitative_examples_full}
\end{table*}

The quantitative distribution shift measures reported in Section~\ref{sec:results_diverge} establish that natural prompt distributions shift significantly over time, between user groups, and across geographic regions, but do not illuminate specific changes in the prompts themselves. To motivate future investigation into the underlying drivers of natural prompt shift, we conduct an initial qualitative analysis of prompt pairs sampled from ID and OOD splits along each axis. For each axis, we randomly select 3 \{ID, OOD\} splits. Within each split, we construct 10 \textbf{shift-representative} prompt pairs by sampling source prompts from the 10\% highest TTE depth score \cite{seegmiller2023statistical} (most source-representative) ID prompts and target prompts from the 10\% lowest TTE depth (least source-representative) OOD prompts. By sampling ID prompts with high TTE depth (indicating they are highly representative of the source distribution $P$) and OOD prompts with low TTE depth (indicating they are poorly represented by $P$ and thus most characteristic of the shift in target distribution $Q$), each prompt pair is constructed to reflect the contrast between typical in-distribution prompts and prompts most emblematic of the distributional shift. This results in 30 pairs per axis and 90 total. We manually label any observable differences, e.g. in topic, register, linguistic complexity, task type, and specificity. We emphasize that this analysis is exploratory, intended to surface hypotheses and motivate future work rather than draw statistically generalizable conclusions. In addition to our examples and discussion in Table \ref{tab:qualitative_examples} in Section \ref{sec:results_trends}, we give three representative examples from each axis in Table \ref{tab:qualitative_examples_full}.

\section{Tabular Results of Prompt Distribution Shift Over Time, Between User Groups, and Across Locations}
\label{app:shift_results}

As discussed in Section \ref{sec:results_diverge}, the LENS framework measures natural prompt distribution shift using four unsupervised measures of distribution shift: MAUVE (lower MAUVE indicates more shift), $\Delta$-score (lower $\Delta$-score indicates more shift), perplexity (higher perplexity indicates more shift), and BLEU (lower BLEU indicates more shift). Here we give aggregated results over all 3 models in each of the 192 \{ID, OOD\} settings across each axis: time, user group, and geographical location. To compare these estimates against a baseline amount of shift which would be expected in random sampling, we also measure shift between $72$ dataset pairs randomly sampled from all English WildChat prompts.

\begin{table*}
  \caption{ID model loss, win, and tie rates against the oracle model, averaged across 192 \{ID, OOD\} settings and three models for a total of 57.6k OOD evaluation prompts. We find that models trained on prompts from specific user groups and geographic regions perform particularly poorly in OOD settings, recording 87\% and 88\% loss rates vs the oracle model trained on OOD data (disjoint from evaluation OOD data). Results are reported as mean (standard deviation). Individual performance for each of the 192 splits can be found in csv format in supplementary materials.}
  \label{app:tab:performance}
  \centering
    \begin{tabular}{lrrrr}\hline
    \textbf{Axis} &\textbf{Loss Rate vs Oracle} &\textbf{Win Rate vs Oracle} &\textbf{Tie Rate vs Oracle} \\\hline
    Time &0.4435 (0.05) &0.3706 (0.05) &0.1586 (0.04) \\
    User Group &0.8723 (0.04) &0.0180 (0.01) &0.0926 (0.03) \\
    Geography &0.8820 (0.05) &0.0103 (0.01) &0.0884 (0.04) \\
    \hline
    \end{tabular}
\end{table*}

As shown in Figure \ref{fig:shift}, and in tabular format in Table \ref{app:tab:shift}, significant natural prompt distribution shifts are observed across all three axes. The geographical location axis exhibits the strongest shift across all 4 shift measures, for example with a mean MAUVE score of just 0.05 compared to 0.96 baseline random, 0.61 user group, and 0.54 time. This indicates that different geographical behavioral patterns are strongly associated with changes in prompt characteristics. While we subset to only consider English prompts in the WildChat dataset, we hypothesize that part of this measurable difference may be due to different English dialects, a well-known problem for LLMs \cite{srirag2025evaluating}. Natural prompt distribution shifts are also strong between time-shifted datasets, e.g. with an average MAUVE score of 0.54 vs the random baseline MAUVE score of 0.96. While user groups based on adoption rate and usage volume exhibit the least natural prompt distribution shift among the three axes which we evaluate, the average MAUVE (0.61), $\Delta$-score (0.47), and BLEU (0.49) scores in this group still indicate measurable distributional shift compared to baseline (0.96, 0.50, 0.57 average MAUVE, $\Delta$-score, and BLEU scores respectively). These results show that prompt distributions encountered by deployed LLMs can vary substantially depending on when the model is used, who is using it, and where they are located. We give aggregated results in Table \ref{app:tab:shift}.

\section{Tabular Results of LLM Performance Degradation Over Time, Between User Groups, and Across Locations}
\label{app:performance_results}

As discussed in Section \ref{sec:results_degrade}, the LENS framework evaluates LLM OOD performance by comparing trained models against oracle models trained on OOD data. We train and evaluate three pairs of ID/oracle models for each of the 192 \{ID, OOD\} settings described in Section \ref{sec:framework_datasets}. Each evaluation is on 100 OOD evaluation prompts (disjoint from oracle training data) for a total of 57.6k evaluation prompts, generating an ID model response and an oracle model response, then using GPT-4o to judge which model response better follows the user's instructions \cite{zhaowildchat}. See Figure \ref{prompt:llmasjudge} for the full LLM-As-Judge evaluation prompt. 

Average model loss, win, and tie rates against the oracle model for each axis can be seen in Figure \ref{app:tab:performance}. We find that models trained on prompts from specific user groups and geographic regions perform particularly poorly in OOD settings, each recording 88\% loss rate vs the oracle model trained on OOD data. We also find that models trained on time-shifted prompts perform poorly on OOD/``future'' prompts, with a 44\% loss rate against the oracle model and only a 37\% win rate. We give aggregated results in Table \ref{app:tab:performance}, and full results across all 192 splits can be found in csv format in supplementary materials.

\section{User Prompt Shift and Model Degradation Across User Group and Geography Axes}
\label{app:ug_geo_results}

\begin{figure*}
    \centering
    \includegraphics[width=\textwidth]{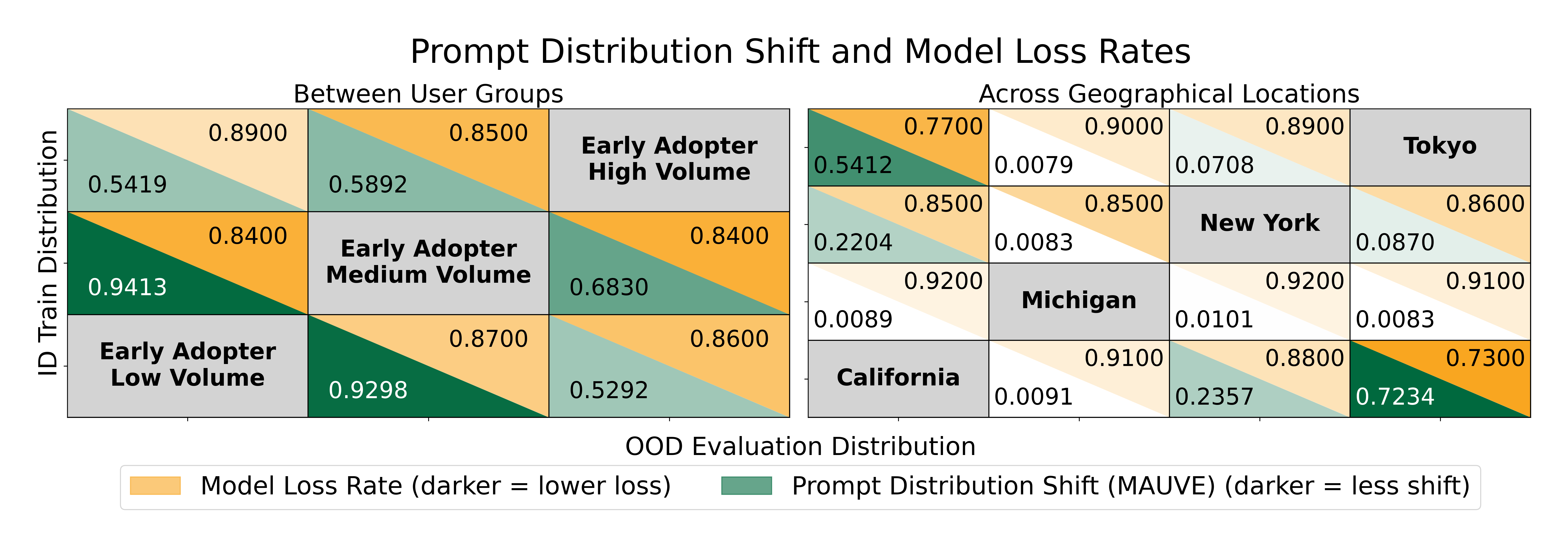}
    \caption{Natural prompt distribution shift (green, with darker meaning less shift) and ID model loss rate (orange, with darker meaning a lower loss rate against the oracle model) are frequently correlated, both between user groups and across geographic regions. For example, users within California and Tokyo exhibit less natural prompt distribution shift than across other regions. Models trained and evaluated on user prompts in these settings also record lower OOD performance loss.  }
    \label{fig:result_2}
\end{figure*}

Figure \ref{fig:result_2} graphically presents selected performance results in the user group and geography settings. In settings where shift is particularly pronounced (low MAUVE), we observe extreme performance drops in OOD settings: ID model loss rate exceeds 88\% on average. In the few cases where MAUVE indicates a lower distributional shift, ID models perform slightly better: for example, the model trained on prompts collected from users in Tokyo, which has a low estimated natural prompt distribution shift from the OOD setting of user prompts from California (MAUVE = 0.54), performs relatively well: 77\% average loss rate as opposed to the 88\% average loss rate. 

\begin{figure*}[t] 
\begin{fullwidthbox}
\textbf{LLM-As-Judge Prompt:} Please act as an impartial judge and evaluate the quality of the responses provided by two AI assistants to the user question displayed below. You should choose the assistant that follows the user's instructions and answers the user's question better. Your evaluation should consider factors such as the helpfulness, relevance, accuracy, depth, creativity, and level of detail of their responses. Begin your evaluation by comparing the two responses and provide a short explanation. Avoid any position biases and ensure that the order in which the responses were presented does not influence your decision. Do not allow the length of the responses to influence your evaluation. Do not favor certain names of the assistants. Be as objective as possible. After providing your explanation, output your final verdict by strictly following this format: ``[[A]]'' if assistant A is better, ``[[B]]'' if assistant B is better, and ``[[C]]'' for a tie.

[User Question]

\{question\}

[The Start of Assistant A's Answer]

\{answer\_a\}

[The End of Assistant A's Answer]

[The Start of Assistant B's Answer]

\{answer\_b\}

[The End of Assistant B's Answer]
\end{fullwidthbox}
\caption{Prompt for comparing ID and oracle model responses to OOD evaluation prompts, following WildChat \cite{zhaowildchat}.}
\label{prompt:llmasjudge}
\end{figure*}
\end{document}